\documentclass[10pt,twocolumn,letterpaper]{article}

\usepackage[pagenumbers]{cvpr}

\usepackage{booktabs}
\usepackage{multirow}
\usepackage{graphicx}
\usepackage{amssymb}
\usepackage{adjustbox}
\usepackage{xcolor,colortbl}
\usepackage{makecell}

\makeatletter
\@ifundefined{Cross}{}{}
\makeatother
\usepackage{marvosym}

\usepackage{cuted}
\usepackage{booktabs}  %
\usepackage{array}     %
\usepackage{longtable} %

\renewcommand{\paragraph}[1]{\vspace{.5em}\noindent\textbf{#1}}

\definecolor{cvprblue}{rgb}{0.21,0.49,0.74}
\definecolor{pinkcolor}{RGB}{230, 0, 115}
\usepackage[pagebackref,breaklinks,colorlinks,allcolors=cvprblue]{hyperref}

\hypersetup{
    colorlinks=true,
    urlcolor=pinkcolor,
}

\title{\raisebox{-0.3\height}{\includegraphics[height=1.8em]{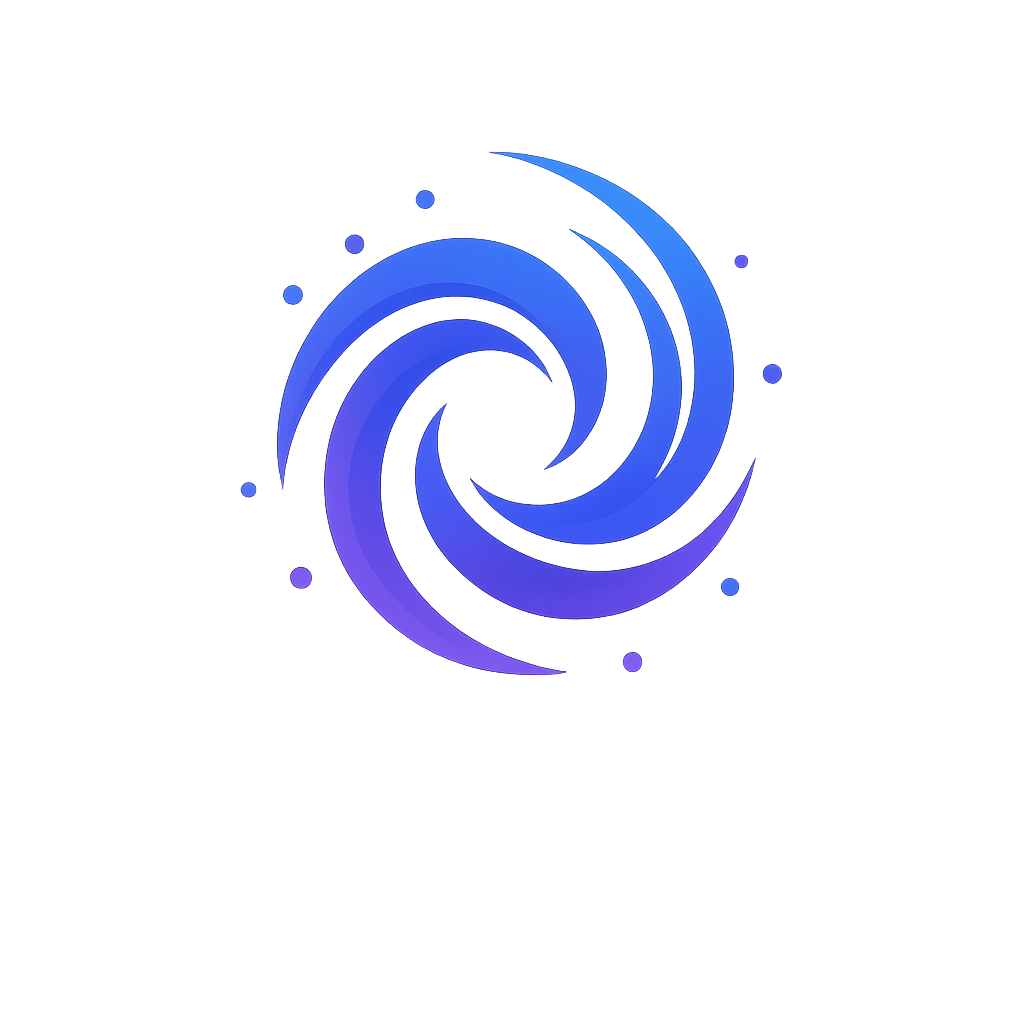}}\textbf{UnityVideo: Unified Multi-Modal Multi-Task Learning \\ for Enhancing World-Aware Video Generation}
\vspace{-0.9em}
}

\author{
Jiehui Huang\textsuperscript{$1,\dagger$} \quad
Yuechen Zhang\textsuperscript{2} \quad
Xu He\textsuperscript{3} \quad
Yuan Gao\textsuperscript{4} \quad
Zhi Cen\textsuperscript{4} \\
Bin Xia\textsuperscript{2} \quad
Yan Zhou\textsuperscript{4} \quad
Xin Tao\textsuperscript{4} \quad
Pengfei Wan\textsuperscript{4} \quad
Jiaya Jia\textsuperscript{1 \Letter} \\[0.3em]
{\textsuperscript{1}HKUST \quad
\textsuperscript{2}CUHK \quad
\textsuperscript{3}Tsinghua University \quad
\textsuperscript{4}Kling Team, Kuaishou Technology} \\[0.3em]
{Projects: \url{https://jackailab.github.io/Projects/UnityVideo}}
}

\newcommand\nnfootnote[1]{%
  \begin{NoHyper}
  \renewcommand\thefootnote{}\footnote{#1}%
  \addtocounter{footnote}{-1}%
  \end{NoHyper}
}

\begin{document}

\twocolumn[{%
\vspace{-0.15in}
\maketitle
\vspace{-0.5in}
\begin{center}
    \captionsetup{type=figure}
    \includegraphics[width=1.0\linewidth]{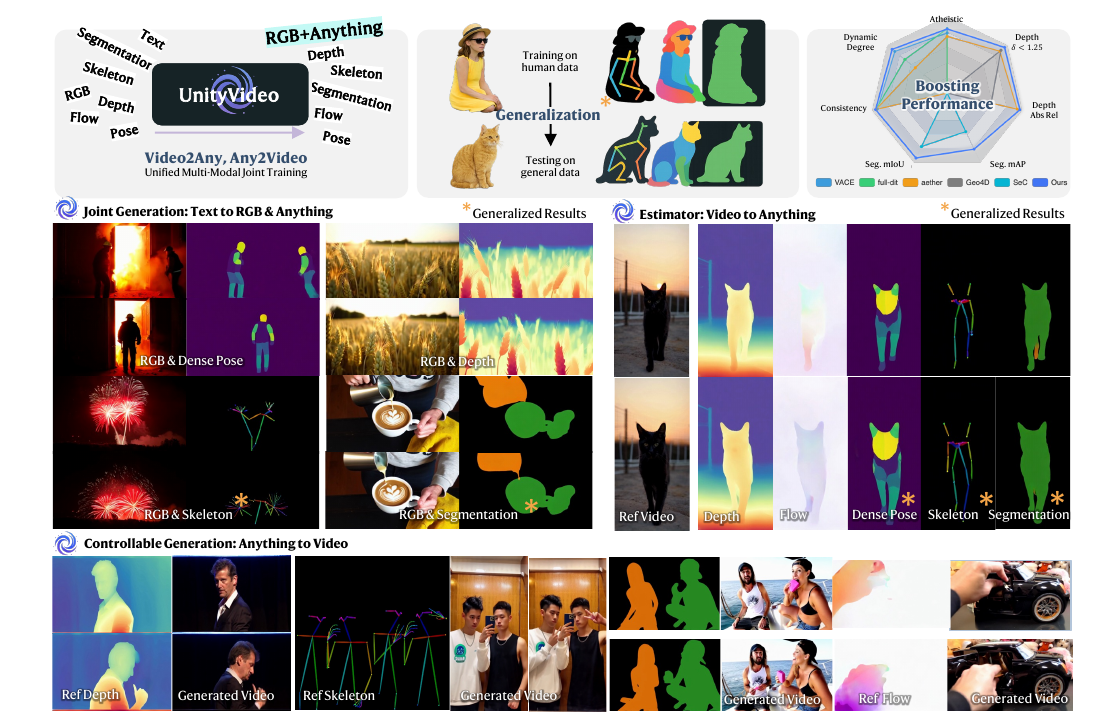}
    \vspace{-0.20in}
    \captionof{figure}{
    In this paper, we present \textbf{UnityVideo}, a unified generalist framework for multi-task multi-modal video understanding. It supports text-to-video generation, controllable generation, and modality estimation, with strong zero-shot generalization to novel objects and styles.
    }
    \label{fig_teaser}
\end{center}
\vspace{-0.05in}
}]

\nnfootnote{
\hspace{-2em}$\dagger$ {This work was conducted during the author's internship at Kling Team, Kuaishou Technology.} \\
\hspace{2em}\Letter~{Corresponding Author.}
}

\vspace{-4mm}
\begin{abstract}
\vspace{-8mm}

\noindent Recent video generation models demonstrate impressive synthesis capabilities but remain limited by single-modality conditioning, constraining their holistic world understanding. This stems from insufficient cross-modal interaction and limited modal diversity for comprehensive world knowledge representation.
To address these limitations, we introduce \textbf{UnityVideo}, a unified framework for world-aware video generation that jointly learns across multiple modalities (segmentation masks, human skeletons, DensePose, optical flow, and depth maps) and training paradigms. Our approach features two core components: (1) dynamic noising to unify heterogeneous training paradigms,
and (2) a modality switcher with an in-context learner that enables unified processing via modular parameters and contextual learning. 
We contribute a large-scale unified dataset with 1.3M samples. Through joint optimization, UnityVideo accelerates convergence and significantly enhances zero-shot generalization to unseen data. We demonstrate that UnityVideo achieves superior video quality, consistency, and improved alignment with physical world constraints.
Code and data can be found at: \url{https://github.com/dvlab-research/UnityVideo}
\end{abstract}

\vspace{-4mm}
\section{Introduction}
\label{sec_intro}
\vspace{-1mm}

Large language models (LLMs) have achieved strong generalization and reasoning ability by unifying diverse text-based modalities, including natural language, code, and mathematical expressions, within a single training paradigm~\cite{lu2024ai, schmidgall2025agent, comanici2025gemini, li2025generation, yang2025qwen3, team2025kimi, liu2024deepseek}. This integration of complementary text sub-modalities improves task performance and supports emergent reasoning. Similarly, recent video foundation models show promising world modeling as scale and parameters increase~\cite{brooks2024video, wan2025wan, kong2024hunyuanvideo, wiedemer2025video, li2024sora}. However, visual scaling has largely centered on RGB video alone, analogous to training language models only on plain text. This gap motivates the question of whether unifying visual sub-modalities--such as depth, optical flow, segmentation, skeleton, and DensePose--can strengthen a model's understanding of the physical world, as unified text learning has done for LLMs.

Recent work indicates that video generation can benefit from single auxiliary input, such as depth maps, optical flow, skeletons, and segmentation masks~\cite{yariv2025through, yan2017skeleton, peng2024controlnext, wang2024motionctrl}. Many approaches use a one-way interaction: conditioning RGB generation on auxiliary modalities for controllable synthesis~\cite{jiang2025vace, ju2025fulldit, wu2025any2caption, cai2025omnivcus}, or predicting these modalities from RGB via inverse estimation~\cite{wang2025vggt, jiang2025geo4d, hu2025depthcrafter}. A few recent frameworks~\cite{team2025aether, baigeovideo, chefer2025videojam, xiu2025egotwin, chen2025deepverse} explore bidirectional interactions and report gains in motion and geometric understanding through shared representations across modalities.

\begin{figure}[t!]
    \centering
    \vspace{-1mm}
    \includegraphics[width=0.98\columnwidth]{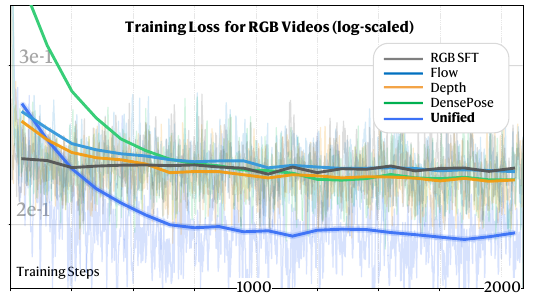}
    \vspace{-2mm}
    \caption{\textbf{Training on unified modalities benefits video generation.} Unified multi-modal and multi-task joint training achieves the lowest final loss on RGB video generation, outperforming single-modality joint training and RGB finetuning baseline.}
    \vspace{-6mm}
    \label{fig_trainingLoss}
\end{figure}

Despite this progress, the effect of a \emph{unified training paradigm} on cross-modal interaction and world awareness remains unclear. Can joint training on multiple modalities and tasks improve reasoning, accelerate convergence, and yield emergent perception? Single-modal learning limits a model’s ability to infer physical dynamics, encouraging distribution fitting rather than reasoning. In practice, different modalities provide complementary cues: instance segmentation separates categories~\cite{koprinska2001temporal, tsai2016video, zou2023segment}, DensePose distinguishes body parts~\cite{guler2018densepose, sanakoyeu2020transferring}, and skeletons encode fine-grained motion~\cite{su2020human, morais2019learning}. {This is demonstrated in~\cref{fig_trainingLoss},} jointly learning from complementary information across diverse modalities {benefits the convergence in video generation}, further offers a path toward more comprehensive world modeling and improved zero-shot generalization.

\textbf{UnityVideo} is presented motivated by these observations. UnityVideo is a unified framework for multimodal video generation, estimation, and joint modeling. UnityVideo integrates multiple modalities and training paradigms to accelerate convergence, enhance zero-shot generalization, and promote mutual gains across tasks and modalities. The framework introduces a light-weight modality-adaptive learner that maps heterogeneous modality signals into a shared feature space, enabling plug-and-play selection of inputs at inference. To further improve generalization, we design an in-context learner that leverages internal contextual cues to enable text-driven video object segmentation without external detectors~\cite{liu2024grounding}. We also devise a dynamic noise scheduling strategy that switches among different training objectives, including joint generation, video estimation, and controllable generation, within a single training cycle to encourage cross-task synergy.

 We curate \textbf{OpenUni}, a large-scale dataset of 1.3M multimodal video samples to enable this unified training paradigm, and construct a high-quality benchmark, \textbf{UniBench}. {UniBench} contains 30K synthetic videos and a subset of the training data, with ground-truth depth and optical flow rendered in Unreal Engine. These resources provide a solid basis for fair and comprehensive evaluation. As shown in Fig~\ref{fig_teaser}, \textit{UnityVideo} is a general-purpose model that performs both video generation and estimation, and it generalizes in a zero-shot manner to novel objects that not provided in training data. Extensive quantitative and qualitative results demonstrate that our model outperforms existing approaches across multiple downstream tasks.
Our main contributions are summarized as follows:

\begin{itemize}
    \item We propose {UnityVideo}, a novel unified framework for integrating multiple video tasks and modalities, enabling mutual knowledge transfer, better convergence, and improved performance over single-task baselines.
    \item We introduce a modality-adaptive switcher, an in-context learner, and a dynamic noise scheduling strategy that together enable efficient joint training across diverse objectives and scalability to larger datasets.
    \item We construct and release {OpenUni}, a 1.3M-pair multimodal video dataset, and {UniBench}, a 30K-sample benchmark derived from Unreal Engine for fair evaluation of unified video models.
\end{itemize}

\section{Releated Work}
\label{sec:related}

\subsection{Video Generation}

Large-scale video generation has advanced world modeling and physical reasoning~\cite{xie2024show, chen2024sharegpt4video, tan2025omni, liu2025javisdit, wiedemer2025video, wang2025universe}, improving a model’s ability to capture physical dynamics~\cite{bansal2025videophy, wiedemer2025video, chen2025towards, ji2025physmaster, lin2025exploring}. 
Recent work integrates additional visual signals such as depth, camera pose, and optical flow to jointly model video~\cite{chefer2025videojam, team2025aether, baigeovideo}. 
Two main directions have emerged: (i) encoding multiple modalities into a shared latent space and using flow-matching to jointly predict video and auxiliary modalities, enabling mutual gains~\cite{chefer2025videojam, xiu2025egotwin}; and (ii) conditioning generation on multi-modal inputs for controllable synthesis, allowing simultaneous compliance with multiple control signals and improved visual quality~\cite{jiang2025vace, ju2025fulldit}. 
Despite strong results, most studies isolate either a single architecture or a single modality, limiting cross-task synergy. 
In contrast, we \emph{unify} multi-task learning in a single framework and analyze how such unification enhances world perception and generalization.

\subsection{Video Reconstruction}

Videos contain rich world knowledge, and classical vision methods estimate depth, camera pose, and optical flow directly from RGB~\cite{krishnan2025orchid, team2025aether}. 
Recent diffusion-based approaches learn bidirectional mappings between conditions and video without external modules, revealing intrinsic bidirectional capacity in flow matching frameworks~\cite{jiang2025geo4d, hu2025depthcrafter, sun2025unigeo}. 
Representative systems such as Aether~\cite{team2025aether}, GeoVideo~\cite{baigeovideo}, and 4DNex~\cite{chen20254dnex} couple video with geometric modalities, and EgoTwin~\cite{xiu2025egotwin} links skeletons and video. 
Bidirectional interactions also appear between video and audio~\cite{wang2025universe} and between video and text~\cite{yan2021videogpt, chen2024sharegpt4video}, for example UniVerse-1 for audio and video~\cite{wang2025universe} and UniVid or Omni-Video for text and video. 
However, prior work has not fully unified diverse modalities or systematically studied their synergy, and it rarely activates in-context abilities for strong zero-shot generalization. 
Our approach addresses these gaps through joint training across modalities and tasks, yielding a unified model with stronger zero-shot performance and clearer insights into cross-modal coupling.

\section{Method}

\begin{figure*}[h!]
    \centering
    \includegraphics[width=1.0\textwidth]{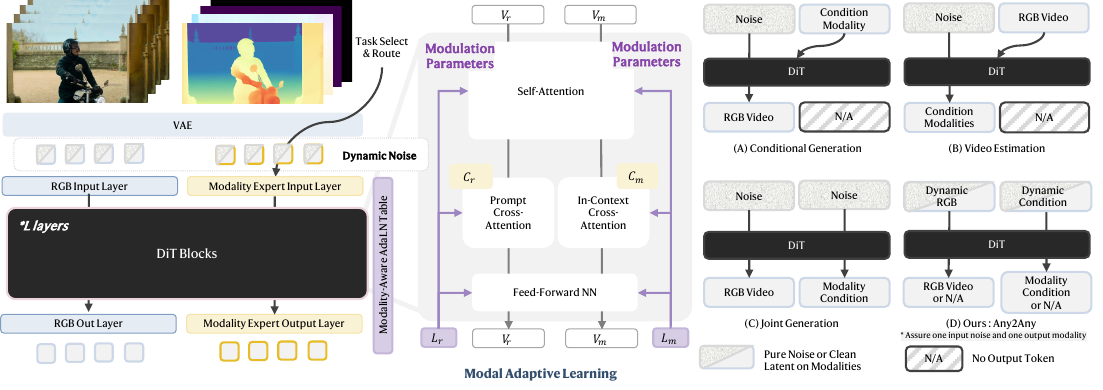}
    \vspace{-5mm}
    \caption{
    \textbf{Overview of UnityVideo.} 
    UnityVideo achieves {task unification} through a dynamic noise injection strategy applied to input tokens (left), and \textit{modality unification} via the proposed Modality-Aware AdaLN Table (center). 
    Specifically, $L_r$ and $L_m$ denote the learnable parameter tables for the RGB modality and auxiliary video-related modalities (e.g., depth, optical flow, DensePose, skeleton), respectively. 
    $C_{r}$ and $C_{m}$ represent the prompt condition for RGB video content and in-context modaliy learning prompt, while $V_r$ and $V_m$ correspond to the token sequences from the RGB and auxiliary modalities, respectively.
    }
    \vspace{-3mm}
    \label{fig_framework}
\end{figure*}

UnityVideo unifies video generation and multimodal understanding within a single diffusion transformer. As illustrated in Fig.~\ref{fig_framework}, the model processes RGB video $V_r$, text condition $C$, and auxiliary modality $V_m$ through a shared DiT backbone $u(\cdot)$. During training, we dynamically sample task types and apply corresponding noise schedules (Sec.~\ref{sec:task_unify_tasks}). To handle multiple modalities within this unified architecture, we introduce an In-Context Learner and a Modality-Adaptive Switcher (Sec.~\ref{sec:task_unify_modalities}). Through progressive curriculum training (Sec.~\ref{sec:training_strategy}), the model achieves simultaneous convergence across all tasks and modalities.

\subsection{Unifying Multiple Tasks}
\label{sec:task_unify_tasks}
Conventional video generation models are trained for specific tasks in isolation, limiting their ability to leverage cross-task knowledge. We extend the flow matching framework~\cite{lipman2022flow} to support three complementary training paradigms within a single architecture.
UnityVideo simultaneously handles three objectives: generating RGB videos from auxiliary modalities ($u(V_r|V_m, C)$), estimating auxiliary modalities from RGB videos ($u(V_m|V_r)$), and jointly generating both from noise ($u(V_r,V_m|C)$).
The $V_r$ and $V_m$ tokens are concatenated along the width dimension and interact through the self-attention module. 
Following~\cite{tan2025ominicontrol, ju2025fulldit}, we incorporate 3D RoPE within the DiT backbone's self-attention to effectively distinguish cross-modal spatiotemporal positions.

\paragraph{Dynamic Task Routing.} To enable concurrent optimization across these three paradigms, we introduce probabilistic task selection during training. At each iteration, we sample one task type with probabilities $p_{\text{cond}}$, $p_{\text{est}}$, and $p_{\text{joint}}$ (where $p_{\text{cond}} + p_{\text{est}} + p_{\text{joint}} = 1$), which determines the noise schedule applied to RGB and modality tokens at timestep $t$. For conditional generation, as depicted in the right part of Fig.~\ref{fig_framework}, RGB tokens are gradually denoised from noise ($t \sim [0,1]$) while modality tokens remain clean ($t=0$). For modality estimation, RGB tokens remain clean while modality tokens are noised. For joint generation, both token types are independently corrupted with noise. We assign task probabilities inversely proportional to their learning difficulty: $p_{\text{cond}} < p_{\text{est}} < p_{\text{joint}}$. This strategy prevents the catastrophic forgetting common in sequential stage-wise training, allowing the model to learn all three distributions concurrently.

\subsection{Unifying Multiple Modalities}
\label{sec:task_unify_modalities}
Joint training across different modalities can significantly enhance the performance of individual tasks, as in~\cref{fig_trainingLoss}. 
However, processing diverse modalities with shared parameters requires explicit mechanisms to distinguish them. 
We introduce two complementary designs: a context learner for semantic-level modality awareness and a modality-adaptive switcher for architecture-level modulation.

\paragraph{In-Context Learner.} To leverage the model's inherent contextual reasoning capability, we inject modality-specific textual prompts $C_m$ that describe the modality type (e.g., ``depth map,'' ``human skeleton'') rather than video content. This design fundamentally differs from content-describing captions $C_r$. Given concatenated RGB tokens $V_r$ and modality tokens $V_m$, we perform dual-branch cross-attention separately: $V_r' = \text{CrossAttn}(V_r, C_r)$ for RGB features with content captions, and $V_m' = \text{CrossAttn}(V_m, C_m)$ for modality features with type descriptions, before recombining them for subsequent processing. This lightweight mechanism introduces negligible computational overhead while enabling compositional generalization. For instance, training with the phrase ``two persons'' allows the model to generalize to ``two objects'' during segmentation tasks, as the model learns to interpret modality-level semantics rather than memorizing content-specific patterns. Detailed analysis is provided in the experimental section.

\paragraph{Modality-Adaptive Switcher.} While text-based differentiation provides semantic awareness, it may become insufficient as the number of modalities scales. We therefore introduce a learnable embedding list $\mathbf{L}_m = \{L_1, L_2, \ldots, L_k\}$ for $k$ modalities to enable explicit architecture-level modulation. Within each DiT block, AdaLN-Zero~\cite{peebles2023scalable} produces modulation parameters (scale $\gamma$, shift $\beta$, gate $\alpha$) for RGB features based on timestep embeddings. We extend this mechanism by learning modality-specific parameters: $\gamma_m, \beta_m, \alpha_m = \text{MLP}(L_m + t_{\text{emb}})$, where $L_m \in P_m$ is the modality embedding and $t_{\text{emb}}$ is the timestep embedding. This design enables plug-and-play modality selection during inference. To further reduce modality confusion and stabilize outputs, we initialize modality expert input-output layers as a dedicated encoding and prediction head for each modality.
Further details are provided in the Appendix~\ref{Appendix_more_analysis_of_model_design}.

\begin{figure}[t]
    \centering
    \includegraphics[width=0.98\columnwidth]{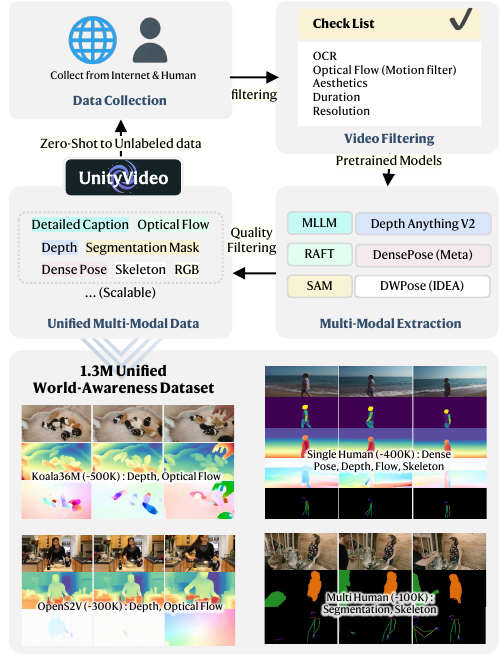}
    \caption{\textbf{OpenUni dataset.} OpenUni contains 1.3M pairs of unified multimodal data, designed to enrich video modalities with more comprehensive world perception.
    }
    \label{fig_data}
    \vspace{-7mm}
\end{figure}

\subsection{Training Strategy}
\label{sec:training_strategy}
\paragraph{Curriculum Learning for Multiple Modalities.} Naively training all modalities jointly from scratch leads to slow convergence and suboptimal performance. We categorize modalities into two groups based on their spatial alignment properties. Pixel-aligned modalities (optical flow, depth, DensePose) allow direct pixel-to-pixel correspondence with RGB frames, while pixel-unaligned modalities (segmentation, skeleton) require additional visual rendering steps. We adopt a two-stage curriculum strategy: Stage 1 trains only pixel-aligned modalities on curated single-person data, establishing a strong foundation for spatial correspondence learning. Stage 2 incorporates all modalities and diverse scene datasets, covering both human-centric and general scenarios. This progressive strategy enables the model to understand all five modalities while supporting robust zero-shot inference on unseen modality combinations.

\paragraph{OpenUni Dataset.} Our training data comprises 1.3 million video clips spanning five modalities: optical flow, depth, dense pose, skeleton, and segmentation. As illustrated in Figure~\ref{fig_data}, we collect real-world videos from multiple sources and extract modality annotations using pre-trained models. The dataset includes 370,358 single-person clips, 97,468 two-person clips, 489,445 clips from Koala36M~\cite{wang2025koala}, and 343,558 clips from OpenS2V~\cite{yuan2025opens2v}, totaling 1.3M samples for training. To prevent overfitting to specific datasets or modalities, we partition each batch into four balanced groups, ensuring uniform sampling across all modalities and sources. More details on training data are provided in Appendix~\ref{Appendix_details_of_OpenUni_and_UniBench}.

\subsection{Training Objective}
Following Conditional Flow Matching~\cite{lipman2022flow}, our framework employs a dynamic training strategy that adaptively switches between three modes by selectively noising different modalities. The mode-specific losses are:
\begin{align}
\mathcal{L}_{\text{cond}}(\theta; t) &= \mathbb{E}\left[\|u_\theta(r_t, [m_0, c_{\text{txt}}], t) - v_r\|^2\right], \label{eq:cond}\\
\mathcal{L}_{\text{est}}(\theta; t) &= \mathbb{E}\left[\|u_\theta(m_t, r_0, t) - v_m\|^2\right], \label{eq:est}\\
\mathcal{L}_{\text{joint}}(\theta; t) &= \mathbb{E}\left[\|u_\theta([r_t, m_t], c_{\text{txt}}, t) - [v_r, v_m]\|^2\right], \label{eq:joint}
\end{align}
where $r_t = (1-t)r_0 + tr_1$ and $m_t = (1-t)m_0 + tm_1$ denote the interpolated latents at timestep $t \in [0,1]$, with $r$ and $m$ representing RGB video and auxiliary modality (e.g., optical flow, depth) respectively. The velocity fields are defined as $v_r = r_1 - r_0$ and $v_m = m_1 - m_0$, where $r_0, m_0$ are clean latents encoded from real data and $r_1, m_1 \sim \mathcal{N}(0, I)$ are independent Gaussian noise. The text conditioning $c_{\text{txt}}$ is obtained from a pre-trained text encoder. Eq.~\eqref{eq:cond} enables conditional generation of RGB video from auxiliary modality, Eq.~\eqref{eq:est} performs modality estimation from RGB video, and Eq.~\eqref{eq:joint} jointly generates both modalities from text.

During training, each sample in a batch is randomly assigned to one of the three modes, enabling all tasks to contribute gradients within a single optimization step. This unified formulation allows seamless multi-task learning within a single architecture.

\section{Experiment}

In this section, we first provide implementation details in Sec.~\ref{implementation_details}, followed by the main results in Sec.~\ref{main_results}. We conduct extensive benchmarks on both modality estimation and video generation tasks, comparing UnityVideo against state-of-the-art methods. The results demonstrate that UnityVideo exhibits strong unified capabilities across all settings. Subsequently, Sec.~\ref{abalation_study} presents ablation studies that validate the effectiveness of our design choices. Finally, we analyze the convergence behavior and zero-shot generalization ability of UnityVideo, complemented by a user study. Additional analysis of UnityVideo's zero-shot generalization and its reasoning abilities about video modalities are provided in the Appendix~\ref{Appendix_more_experiments_and_analysis}.

\begin{table*}[t]
\centering
\caption{
Quantitative comparison of UnityVideo on \colorbox{gray!15}{controllable generation}, \colorbox{cyan!15}{text-to-video generation}, and \colorbox{yellow!15}{video estimation} tasks. Best results are in \textbf{bold}, and second-best results are \underline{underlined}. Compared to state-of-the-art methods and commercial models, UnityVideo achieves superior or competitive performance across all metrics.
}
\vspace{-2mm}
{
\scriptsize
\label{tab_comprehensive_comparison}
\resizebox{\textwidth}{!}{
\begin{tabular}{ll|cccc|cc|cc}
\toprule
\multicolumn{2}{c|}{} & \multicolumn{4}{c|}{\textbf{Video Generation - VBench \& UniBench Dataset}} & \multicolumn{4}{c}{\textbf{Video Estimation - UniBench Dataset}} \\
\cmidrule(lr){3-6} \cmidrule(lr){7-10}
\multicolumn{2}{c|}{} & \multicolumn{4}{c|}{\textbf{VBench}} & \multicolumn{2}{c|}{\textbf{Segmentation}} & \multicolumn{2}{c}{\textbf{Depth}} \\
\cmidrule(lr){3-6} \cmidrule(lr){7-8} \cmidrule(lr){9-10}
\textbf{Tasks} & \textbf{Models} & \textbf{Background} & \textbf{Aesthetic} & \textbf{Overall} & \textbf{Dynamic} & \textbf{mIoU}  $\uparrow$ & \textbf{mAP}  $\uparrow$ & \textbf{Abs Rel} $\downarrow$ & $\boldsymbol{\delta < 1.25}$ $\uparrow$ \\
& & \textbf{Consistency} & \textbf{Quality} & \textbf{Consistency} & \textbf{Degree} & & & & \\
\midrule
\multirow{5}{*}{Text2Video} 
& Kling1.6 & 95.33 & 60.48 & 21.76 & \underline{47.05} & - & - & - & - \\
& OpenSora2 & 96.51 & 61.51 & {19.87} & 34.48 & - & - & - & - \\
& HunyuanVideo-13B & 96.28 & 53.45 & \underline{22.61} & 41.18 & - & - & - & - \\
& Wan2.1-14B & \underline{96.78} & \underline{63.66} & 21.53 & 34.31 & - & - & - & - \\
& Aether & 95.28 & 48.25 & 20.26 & 37.32 & - & - & \underline{0.025} & \underline{97.95} \\
\midrule
\multirow{2}{*}{\begin{tabular}[c]{@{}l@{}}Controllable\\Generation\end{tabular}} 
& full-dit & \underline{95.58} & \textbf{54.82} & \underline{20.12} & 49.50 & - & - & - & - \\
& VACE & {93.61} & 51.24 & 17.52 & \underline{61.32} & - & - & - & - \\
\midrule
\multirow{2}{*}{\begin{tabular}[c]{@{}l@{}}Depth Video\\Reconstruction\end{tabular}} 
& depth-crafter & - & - & - & - & - & - & 0.065 & 96.94 \\
& Geo4D & - & - & - & - & - & - & 0.053 & {97.94} \\
\midrule
\multirow{2}{*}{\begin{tabular}[c]{@{}l@{}}Video\\Segmentation\end{tabular}} 
& SAMWISE & - & - & - & - & 62.21 & 20.12 & - & - \\
& SeC & - & - & - & - & \underline{65.52} & \underline{22.23} & - & - \\
\midrule
\rowcolor{gray!15} 
Unified ControGen, & UnityVideo & \textbf{96.04} & \underline{54.63} & \textbf{21.86} & \textbf{64.42} & \cellcolor{yellow!15} & \cellcolor{yellow!15} & \cellcolor{yellow!15} & \cellcolor{yellow!15} \\
\rowcolor{cyan!15} 
T2V, and Estimation & UnityVideo & \textbf{97.44} & \textbf{64.12} & \textbf{23.57} & \textbf{47.76} & \cellcolor{yellow!15}\textbf{68.82} & \cellcolor{yellow!15}\textbf{23.25} & \cellcolor{yellow!15}\textbf{0.022} & \cellcolor{yellow!15}\textbf{98.98} \\
\bottomrule
\end{tabular}
}
\vspace{-2mm}
}
\end{table*}

\subsection{Experimental Setup}
\label{implementation_details}

\paragraph{Training Details.}  
We use an internal DiT backbone with 10B parameters as our core architecture. Training is conducted in two stages. In the first stage, the model is trained on a human-centric dataset containing 500K video clips for 16K steps. In the second stage, we scale up training to a larger dataset of 1.3M video clips for an additional 40K steps. The model is trained with a batch size of 32 and a learning rate of $5\times10^{-5}$. During inference, we use 50 DDIM sampling steps with a CFG scale of 7.5.

\paragraph{Baselines.}  
Since our framework introduces a novel unified paradigm for video generation and estimation, no directly comparable models exist. We therefore evaluate against leading models in three related categories: 
(1) \textbf{Video Generation:} We compare with text-to-video models, including the commercial model {Keling-1.6}, and open-source models {OpenSora}~\cite{peng2025open}, {Hunyuan-13B}~\cite{kong2024hunyuanvideo}, and {Wan-2.1-13B}~\cite{wan2025wan}. For controllable generation, we include {VACE}~\cite{jiang2025vace} and {Full-DiT}~\cite{ju2025fulldit}. We also consider models capable of jointly generating video and depth, such as {Aether}~\cite{team2025aether}.
(2) \textbf{Video Estimation:} We evaluate against diffusion-based depth estimation models, including {DepthCrafter}~\cite{hu2025depthcrafter}, {Geo4D}~\cite{jiang2025geo4d}, and {Aether}~\cite{team2025aether}. Additional results are provided in the Appendix.  
(3) \textbf{Video Segmentation:} We compare with two recent segmentation models that support prompt-based object segmentation, {SAMWISE}~\cite{cuttano2025samwise} and {SeC}~\cite{zhang2025sec}.  

To ensure fair comparisons, all evaluations are conducted on the public VBench~\cite{huang2024vbench} dataset and our newly constructed UniBench dataset, specifically designed for unified video tasks. UniBench comprises 200 high-quality samples obtained from Unreal Engine (UE) for accurate video estimation evaluation~\cite{yang2024depth}, and 200 manually curated samples covering diverse modalities from real-world videos~\cite{ju2025fulldit} for controllable generation and segmentation assessments. More details are provided in Appendix~\ref{Appendix_details_of_OpenUni_and_UniBench}.

\paragraph{Evaluation Metrics.}  
To comprehensively assess the performance of our model, we evaluate it across three categories of metrics.  
(1) \textbf{Video Quality.} We measure visual and temporal quality using multiple perceptual and consistency-based metrics~\cite{huang2024vbench}, including {subject consistency}, {background consistency}, {aesthetic quality}, {imaging quality}, {temporal flickering}, {motion smoothness}, and {dynamic degree}. These metrics collectively evaluate spatial fidelity, aesthetic appeal, and temporal coherence of the generated videos.
(2) \textbf{Depth Estimation.} For quantitative evaluation of video-based depth prediction~\cite{hu2025depthcrafter}, we report the {absolute relative error (AbsRel)} and the percentage of predicted depths within a 1.25 factor of the ground truth ($\delta < 1.25$).
(3) \textbf{Video Segmentation.} To evaluate segmentation accuracy~\cite{zou2023segment, zhang2025sec}, we adopt standard metrics for both instance and semantic segmentation tasks, namely the {mean Average Precision (mAP)} and {mean Intersection-over-Union (mIoU)}.

\begin{figure*}[h]
    \centering
    \includegraphics[width=1.0\linewidth]{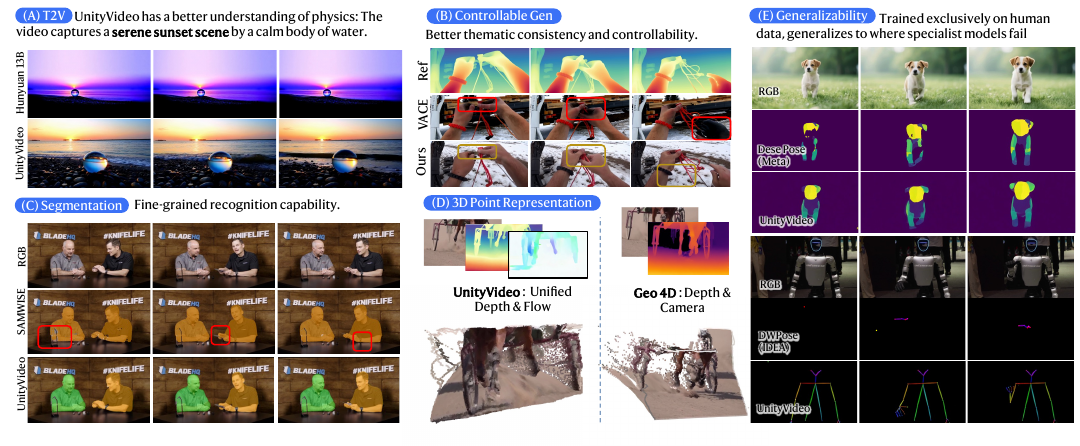}
    \vspace{-8mm}
    \caption{
    \textbf{Comparison with state-of-the-art methods across diverse tasks}. UnityVideo exhibits superior physical reasoning, better adherence to control conditions, and a more detailed understanding of auxiliary modalities. 
 }
 \vspace{-4mm}
    \label{compare_all_base}
\end{figure*}

\subsection{Main Results}
\label{main_results}

This section validates the superior performance of UnityVideo compared to single-task approaches. We comprehensively evaluate UnityVideo on text-to-video generation, controllable generation, and modality estimation tasks, demonstrating both improved generation quality and enhanced world perception capabilities.

\paragraph{Quantitative Comparison.} As shown in~\cref{tab_comprehensive_comparison}, {UnityVideo} achieves competitive results across all tasks, demonstrating strong overall performance. For text-to-video generation, we report the result of the depth-RGB joint generation. Our model obtains the best results on all metrics. We attribute this to joint training across multiple modalities, which enables collaborative refinement and enhances the model's world perception capabilities, leading to superior video quality.

Compared to previous controllable generation methods, {UnityVideo} excels in {background consistency}, {overall consistency}, and {dynamic degree}, while maintaining competitive {aesthetic quality}. This indicates that our model better understands and leverages control conditions, benefiting from multi-task joint training that enables the model to go beyond simply following control signals.
Furthermore, through joint training with multimodal data, {UnityVideo} outperforms single-modality models such as {Geo4D}, {Aether}, and {SeC} on both video segmentation and depth estimation tasks. These results confirm that the unified training framework enhances the model's perception and reasoning capabilities for complex visual scenes.

\paragraph{Qualitative Comparison.}
As shown in Fig.~\ref{compare_all_base} (A), compared to advanced text-to-video models, {UnityVideo} demonstrates superior world perception. Our model exhibits stronger adherence to physical principles, more accurately reflecting the physical phenomenon of refraction. Furthermore, as shown in Fig.~\ref{compare_all_base} (B), compared to advanced controllable generation methods, {UnityVideo} not only follows depth guidance more faithfully but also maintains overall video quality. In contrast, other methods often exhibit noticeable background flickering, with subject regions sometimes distorted by surrounding context.

For modality estimation tasks, as shown in Fig.~\ref{compare_all_base} (C) and (D), {UnityVideo} produces finer edge details,a  wider field of view, and accurate 3D point clouds, benefiting from the complementary nature of multiple modalities. Similarly, in other modality estimation tasks (Fig.~\ref{compare_all_base} (E)), our model demonstrates strong reasoning capabilities, achieving accurate estimation on unseen data and overcoming the overfitting issues observed in other specialized models~\cite{guler2018densepose,yang2023effective}.

Overall, these qualitative results confirm that joint training across multiple tasks and modalities yields significant improvements over single-task or single-modality approaches. This unified framework proves effective in enhancing the model's perception and reasoning capabilities for the physical world. Additional visual results can be found in Appendix~\ref{Appendix_more_visuals_and_applications}.

\subsection{Abalation Study}
Our ablation study addresses two core questions: (a) Does unified training across multiple modalities and tasks enable mutual benefits between modalities, and in what aspects are these benefits manifested? (b) Are our proposed architectural designs effective? What roles do the In-Context Learner and Modality Switcher play in the model? The following experiments address these questions.

\label{abalation_study}
\paragraph{Impact of Different Modalities.}
To quantitatively evaluate the impact of unified multimodal training on video generation, we compare two commonly used modalities, {depth} and {optical flow}, as shown in~\cref{tab_comparison_modalities}. The results show that joint training consistently improves performance across all metrics compared to the baseline. Furthermore, unified training with multiple modalities yields additional gains, particularly in {image quality} and {overall consistency}. This indicates that unifying diverse modalities not only provides complementary supervision during training but also enables mutual enhancement between modalities.

\begin{table}[t]
\centering
{\scriptsize
\caption{
Ablation study comparing single-modality and multimodal training. \textit{Only}: single modality; \textit{Ours}: multiple modalities.
}
\vspace{-2mm}
\label{tab_comparison_modalities}
\resizebox{1\linewidth}{!}{
\begin{tabular}{l|cccc}
\toprule
& \textbf{Subject} & \textbf{Background} & \textbf{Imaging} & \textbf{Overall} \\
& \textbf{Consistency} & \textbf{Consistency} & \textbf{Quality} & \textbf{Consistency} \\
\midrule
Baseline & 96.51 & 96.06 & 64.99 & 23.17 \\
Only Flow & 97.82 & {97.14} & 67.34 & 23.70 \\
Only Depth & \textbf{98.13} & \textbf{97.29} & 69.09 & 23.48 \\
\rowcolor{gray!15} \textbf{Ours}-Flow & 97.97 (+1.46) & {97.19} (+1.13) & \textbf{69.36} (+4.37) & \underline{23.74} (+0.57) \\
\rowcolor{gray!15} \textbf{Ours}-Depth & \underline{98.01} (+1.50) & \underline{97.24} (+1.18) & \underline{69.18} (+4.19) & \textbf{23.75} (+0.58) \\
\bottomrule
\end{tabular}
}
\vspace{-1mm}
}
\end{table}

\begin{table}[t]
\centering
\caption{
Ablation study on single-task versus unified multi-task training. \textit{Only}: single-task; \textit{Ours}: unified multi-task.
}
\vspace{-2mm}
\label{tab_comparison_tasks}
{\scriptsize
\resizebox{1\linewidth}{!}{
\begin{tabular}{l|cccc}
\toprule
& \textbf{Subject} & \textbf{Background} & \textbf{Temporal} & \textbf{Motion} \\
& \textbf{Consistency} & \textbf{Consistency} & \textbf{Flickering} & \textbf{Smoothness} \\
\midrule
Baseline & 96.51 & 96.06 & 98.73 & 99.30 \\
Only ControlGen & 96.53 & {95.58} & 98.45 & 99.28 \\
Only JointGen & \textbf{98.01} & \textbf{97.24} & \underline{99.10} & \underline{99.44} \\
\rowcolor{gray!15} \textbf{Ours}-ControlGen & 96.53 (+0.02) & {96.08 (+0.02)} & {98.79 (+0.06)} & {99.38 (+0.08)} \\
\rowcolor{gray!15} \textbf{Ours}-JointGen & \underline{97.94 (+1.43)} & \underline{97.18 (+0.63)} & \textbf{99.13 (+0.40)}  & \textbf{99.48 (+0.18)} \\
\bottomrule
\end{tabular}
}
}
\vspace{-5mm}
\end{table}

\paragraph{Effect of Multi-Task Training.}
To further quantify the mutual benefits between different training tasks within our unified framework, we train models separately on {Joint Generation} and {Controllable Generation} tasks, both guided by depth modality. Results are summarized in~\cref{tab_comparison_tasks}. We find that training only on the ControlGen task leads to performance degradation compared to the baseline. However, unified multi-task training recovers and even surpasses this performance, achieving improvements across all metrics. Similarly, compared to training only on Joint Generation, unified training shows only slight decreases in {subject consistency} and {background consistency}, while overall performance still outperforms the baseline, demonstrating the effectiveness of multi-task interaction.

\begin{figure}[t]
    \centering
    \includegraphics[width=0.98\columnwidth]{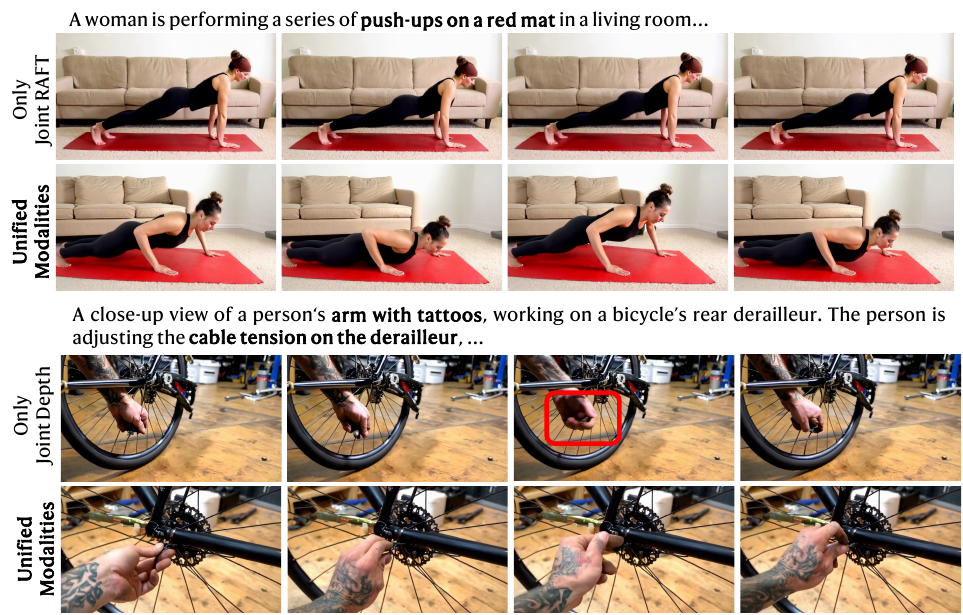}
    \caption{
    Unlike single-modality training, \textbf{unified multimodal} learning provides complementary supervision that strengthens both motion understanding and geometric perception.}
    \vspace{-4mm}
    \label{aba_compare_modalities}
\end{figure}

\paragraph{Impact of Architectural Design.}
We investigate the impact of two architectural strategies, {In-Context Learner} and {Modality Switcher}, on model performance. To ensure consistent evaluation, we perform text-to-video generation conditioned on depth guidance during inference. Results shown in~\cref{tab_comparison_structures} and~\cref{aba_compare_modalities} demonstrate that each strategy effectively improves performance through multimodal fusion. Furthermore, combining both strategies yields additional significant gains, confirming their complementary roles in facilitating unified multimodal learning.

\begin{table}[t]
\centering
\caption{
Ablation study on architectural designs.
}
\vspace{-2mm}
{
\scriptsize
\label{tab_comparison_structures}
\resizebox{1\linewidth}{!}{
\begin{tabular}{l|cccc}
\toprule
& \textbf{Subject} & \textbf{Background} & \textbf{Temporal} & \textbf{Motion} \\
& \textbf{Consistency} & \textbf{Consistency} & \textbf{Flickering} & \textbf{Smoothness} \\
\midrule
Baseline & {96.51} & {96.06} & 98.73 & 99.30 \\
w/ In-Context Learner & {97.92} & {97.08} & 99.04 & 99.42 \\
w/ Modality Switcher & \underline{97.94} & \underline{97.18} & \underline{99.13} & \underline{99.48} \\
Ours & \textbf{98.31} & \textbf{97.54} & \textbf{99.35} & \textbf{99.54} \\
\bottomrule
\end{tabular}
}
}
\end{table}

\begin{table}[t]
\vspace{-2mm}
\centering
\caption{
World perception evaluation comparing UnityVideo with state-of-the-art models.
}
\vspace{-1mm}
{
\scriptsize
\label{tab_user_study}
\resizebox{1\linewidth}{!}{
\begin{tabular}{l|ccc|cc}
\toprule
& \multicolumn{3}{c|}{\textbf{User Study Score (\%)}} & \multicolumn{2}{c}{\textbf{Automatic Score}} \\
\cmidrule(lr){2-4} \cmidrule(lr){5-6}
& Physical & Semantic & Overall & Subject & Motion \\
& Quality & Quality & Preference & Consistency & Smoothness \\
\midrule
Kling1.6 & 10.15 & 21.25 & 20.20 & 83.47 & {98.08} \\
HunyuanVideo & 24.15 & \underline{26.10} & 20.35 & 97.53 & \underline{98.35} \\
Wan2.1 & \underline{27.20} & 22.40 & \underline{27.65} & \underline{97.73} & 98.30 \\
\midrule
Ours & \textbf{38.50} & \textbf{30.25} & \textbf{31.80} & \textbf{98.01} & \textbf{99.33} \\
\bottomrule
\end{tabular}
}
}
\vspace{-3mm}
\end{table}

\subsection{Model Analyze}
\label{model_analyze}
\begin{figure}[h]
    \centering
    \vspace{-3mm}
    \includegraphics[width=1\linewidth]{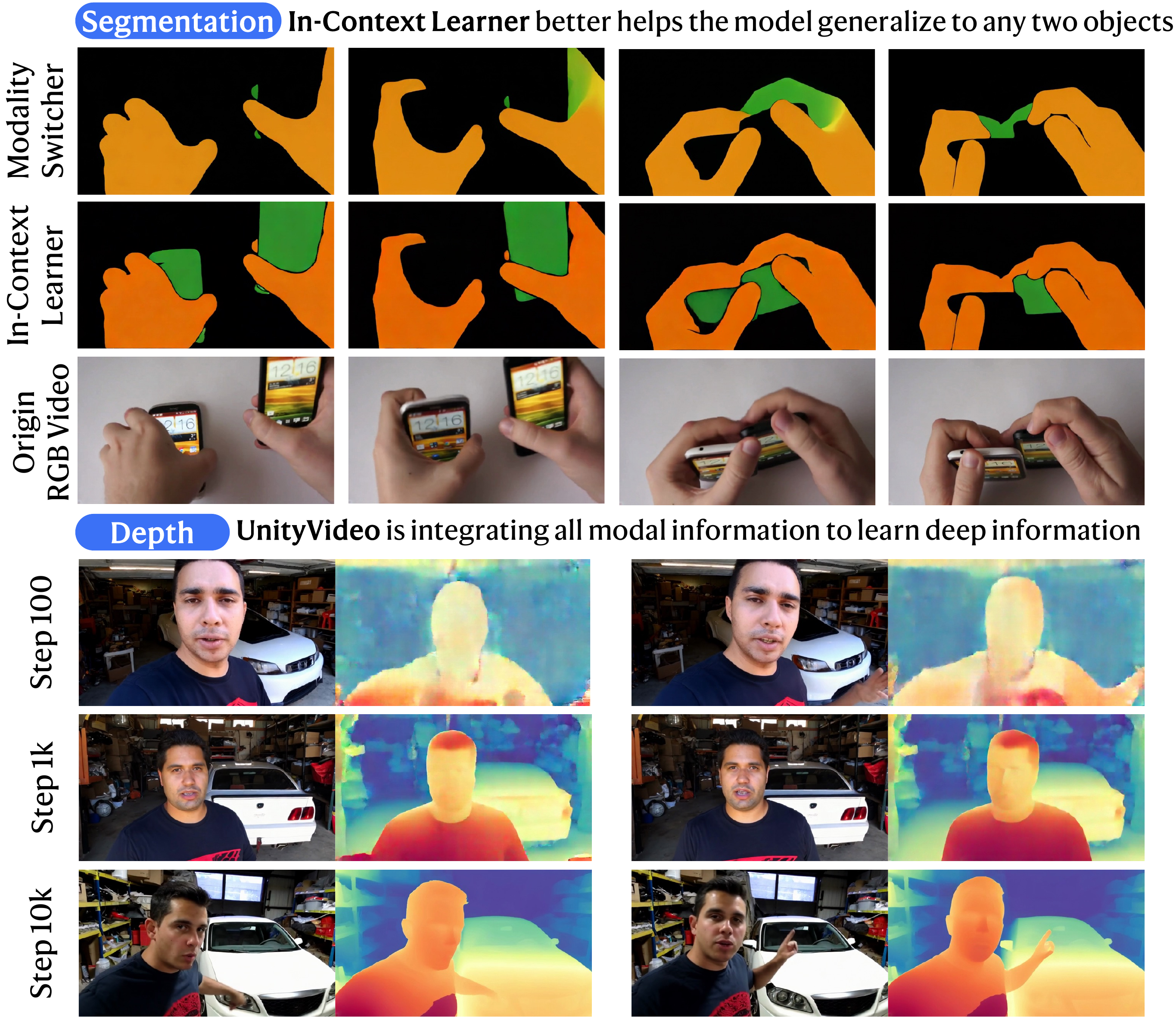}
    \caption{
    The In-Context Learner generalizes segmentation to unseen objects, while unified training enhances depth and semantic understanding in RGB video.
    }
    \vspace{-4mm}
    \label{analysis_converaged}
\end{figure}

As shown in Fig.~\ref{analysis_converaged}, the proposed In-Context Learner effectively generalizes a fixed two-person segmentation task to unseen two-object scenarios. 
In contrast, using only the Modality Switcher fails to achieve such generalization. 
Moreover, during unified training, as the model gradually learns additional modalities (e.g., depth), 
we observe improved motion understanding and more accurate text responses in RGB videos, 
demonstrating the complementary roles of different modalities throughout training.

\paragraph{User Study.}
We conduct a user study using a standard win-rate protocol to evaluate our model’s understanding of the physical world~\cite{wu2025omnigen2}. 
The questionnaire contains 12 randomly selected videos generated with WISA-80K prompts~\cite{wang2025wisa}, presented in random order. 
Each sample is rated by at least three annotators on (i) physical quality, (ii) semantic quality (PF), and (iii) overall quality. 
For automatic evaluation, we adopt two VBench~\cite{huang2024vbench} metrics: dynamism and aesthetics. 
In total, we collect 70 completed responses, and the results are summarized in~\cref{tab_user_study}. 
The study shows that our method achieves the best performance across both human evaluations and automatic metrics.

\section{Limitation and Future Work}
While UnityVideo significantly advances unified visual modeling, several directions remain for future work. The current VAE occasionally introduces reconstruction artifacts, which could be addressed through fine-tuning or improved autoencoder architectures. Additionally, scaling to larger backbones and incorporating more visual modalities may further enhance emergent world understanding capabilities. Despite these limitations, UnityVideo establishes a strong foundation for unified multi-modal video understanding and represents an important step toward comprehensive world models across diverse visual representations.

\section{Conclusion}
\label{sec:conclusion}
We present \textbf{UnityVideo}, a unified framework that models multiple visual modalities and tasks within a single diffusion transformer. By leveraging modal-adaptive learning, UnityVideo enables bidirectional learning between RGB video and auxiliary modalities (depth, optical flow, segmentation, skeleton, and DensePose), achieving mutual enhancement across both tasks. Our experiments demonstrate state-of-the-art performance across diverse benchmarks with strong zero-shot generalization to unseen modality combinations. To support this research, we contribute \textbf{OpenUni}, a large multimodal dataset with 1.3M synchronized samples, and \textbf{UniBench}, a high-quality evaluation benchmark with ground-truth annotations. UnityVideo paves the way toward unified multimodal modeling as a promising step toward next-generation world models.

{
    \small
    \bibliographystyle{ieeenat_fullname}
    \bibliography{main}
}

\clearpage
\clearpage
\setcounter{page}{1}

\maketitlesupplementary

\appendix

\setcounter{figure}{0}
\setcounter{table}{0}

\section*{Appendix}
\addcontentsline{toc}{section}{Appendix}
The appendix contains the following sections:
\begin{itemize}
    \item \hyperref[Appendix_more_analysis_of_model_design]{More Analysis of Model Design}
    \item \hyperref[Appendix_more_experiments_and_analysis]{More Experiments and Analysis}
    \item \hyperref[Appendix_details_of_OpenUni_and_UniBench]{Details of OpenUni and UniBench}
    \item \hyperref[Appendix_more_visuals_and_applications]{More Visuals and Applications}
\end{itemize}

\begin{figure*}[h]
    \centering
    \includegraphics[width=1.0\linewidth]{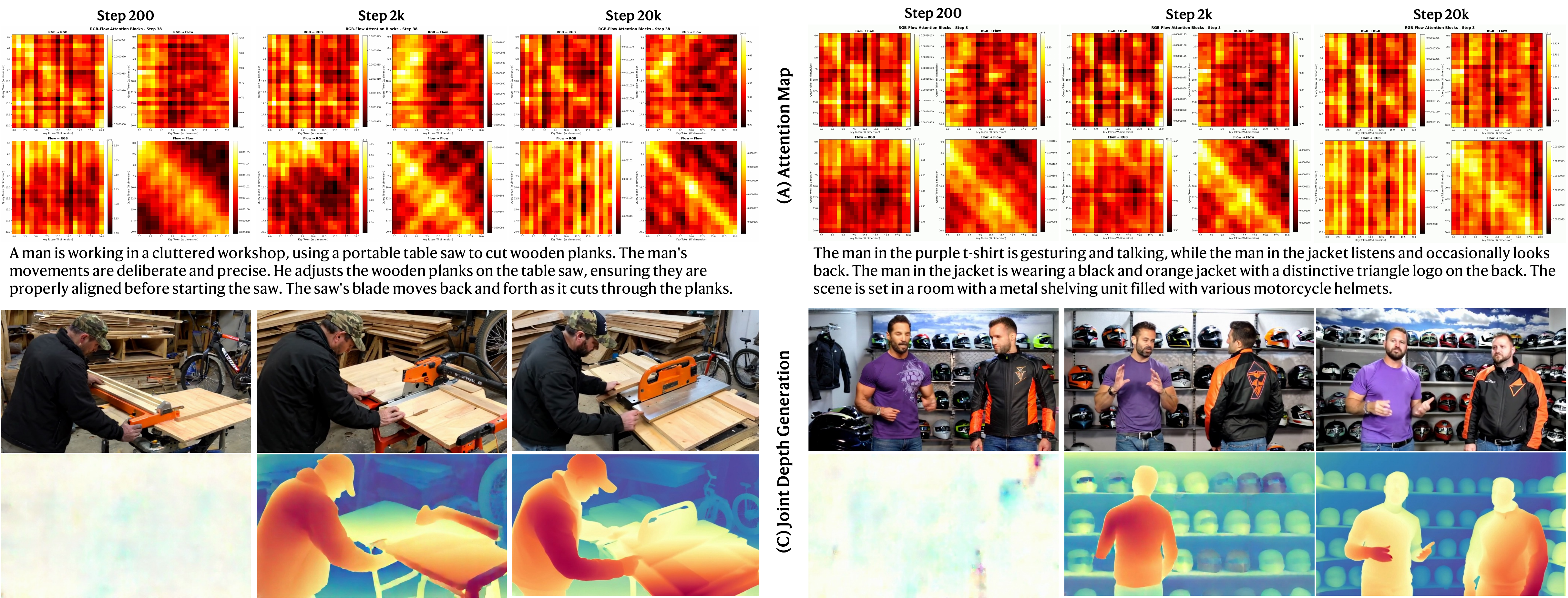}
    \caption{
    \textbf{Evolution of attention patterns in UnityVideo.}  
Analysis of attention maps shows that interactions between RGB and auxiliary modalities strengthen progressively across layers.  
Meanwhile, the model’s text-following behavior and spatial reasoning capabilities also improve, reflecting more coherent cross-modal integration.
 }
 \vspace{-4mm}
    \label{A_analysis_attn_map}
\end{figure*}

\section{More Analysis of Model Design}
\label{Appendix_more_analysis_of_model_design}
\subsection{Modal Interaction Analysis}
To further investigate the cross-modal interactions within our unified framework, we visualize the evolution of self-attention maps throughout the training process. We partition the attention map into four distinct regions based on modality interactions: self-modality regions comprising (RGB, RGB) and (Flow, Flow), and cross-modality regions consisting of (RGB, Flow) and (Flow, RGB), where Flow represents various auxiliary modality features.
As illustrated in Figure~\ref{A_analysis_attn_map}, our analysis reveals three key findings. First, as joint training progresses, the interaction between RGB and auxiliary modalities becomes progressively more pronounced (A), indicating deepening cross-modal feature exchange.
Second, the visualization results demonstrate that the model learns increasingly rich geometric representations with improved text-following capabilities (B), validating the effectiveness of our unified training paradigm in enhancing both visual understanding and conditional generation quality.
This empirical evidence confirms that our unified framework not only enables technical integration of multiple modalities but also facilitates meaningful feature-level interactions that contribute to improved world modeling capabilities.

\subsection{Modality-Specific Output Layers}
While our modality switcher and in-context learner effectively differentiate between modalities, we observed occasional modality confusion as the number of modalities scales. For instance, when instructed to generate segmentation masks, the model infrequently produces skeleton outputs instead. This confusion stems from all modalities sharing a common output layer, which can conflate distinct modality-specific features at the final projection stage.

To address this limitation, we introduce modality-specific output layers (adaptive layer) while maintaining a unified input layer (share layer) for cross-modal information sharing. Each modality receives its own dedicated output projection layer, initialized independently, while the input processing remains shared to preserve inter-modal knowledge transfer. This architectural refinement ensures clear modality boundaries during generation without sacrificing the benefits of unified representation learning.

As shown in Table~\ref{tab_layer_comparison}, this lightweight design effectively eliminates modality confusion during scaled training while maintaining comparable performance across metrics. The modality-specific output layers provide improved flexibility and achieve balanced performance across diverse evaluation criteria, validating this architectural choice for scalable multi-modal generation.

\begin{table}[t]
\centering
\caption{
Comparison of different layer strategies.
}
\vspace{-2mm}
{
\scriptsize
\label{tab_layer_comparison}
\resizebox{1\linewidth}{!}{
\begin{tabular}{l|ccccc}
\toprule
& \textbf{Subject} & \textbf{Background} & \textbf{Temporal} & \textbf{Motion} & \textbf{Averaged} \\
& \textbf{Consistency} & \textbf{Consistency} & \textbf{Flickering} & \textbf{Smoothness} & \\
\midrule
Baseline & 96.51 & 96.06 & 98.73 & 99.30 & 97.650 \\
Share Layer & \textbf{98.31} & \textbf{97.54} & \underline{99.35} & \underline{99.54} & \underline{98.685} \\
Adaptive Layer & \underline{98.26} & \underline{97.49} & \textbf{99.44} & \textbf{99.61} & \textbf{98.700} \\
\bottomrule
\end{tabular}
}
}
\end{table}

\section{More Experiments and Analysis}
\label{Appendix_more_experiments_and_analysis}

\begin{table}[t]
\centering
\caption{
Comparison with standalone T2V. Joint generation achieves better performance, with unified modality showing further improvements.
}
\vspace{-2mm}
{
\scriptsize
\label{tab_comparison_modalities}
\resizebox{1\linewidth}{!}{
\begin{tabular}{l|ccccc}
\toprule
& \textbf{Subject} & \textbf{Background} & \textbf{Imaging} & \textbf{Overall} & \textbf{Averaged} \\
& \textbf{Consistency} & \textbf{Consistency} & \textbf{Quality} & \textbf{Consistency} & \\
\midrule
Baseline & 96.51 & 96.06 & 64.99 & 23.17 & 70.1825 \\
T2V & 96.51 & 97.23 & 66.52 & 23.44 & 70.9250 \\
\midrule
\multicolumn{6}{c}{\textit{Depth Modality}} \\
\midrule
JointGen & \textbf{98.13} & \textbf{97.29} & 69.09 & 23.48 & 71.998 \\
(Depth) & (+1.62) & (+0.06) & (+2.57) & (+0.04) & (+1.073) \\
\cmidrule{1-6}
JointGen & 98.01 & 97.24 & \textbf{69.18} & \textbf{23.75} & \textbf{72.045} \\
(Unified) & (+1.50) & (+0.01) & (+2.66) & (+0.31) & (+1.120) \\
\midrule
\multicolumn{6}{c}{\textit{Optical Flow Modality}} \\
\midrule
JointGen & 97.82 & {97.14} & 67.34 & 23.70 & 71.500 \\
(Optical Flow) & (+1.31) & (-0.09) & (+0.82) & (+0.26) & (+0.575) \\
\cmidrule{1-6}
JointGen & \textbf{97.97} & {97.19} & \textbf{69.36} & \textbf{23.74} & \textbf{72.065} \\
(Unified) & (+1.46) & (-0.04) & (+2.84) & (+0.30) & (+1.140) \\
\midrule
\multicolumn{6}{c}{\textit{Densepose Modality}} \\
\midrule
JointGen & \textbf{98.08} & \textbf{97.38} & 67.05 & 23.49 & 71.500 \\
(Densepose) & (+1.57) & (+0.15) & (+0.53) & (+0.05) & (+0.575) \\
\cmidrule{1-6}
JointGen & 98.03 & 97.30 & \textbf{70.20} & \textbf{23.53} & \textbf{72.265} \\
(Unified) & (+1.52) & (+0.07) & (+3.68) & (+0.09) & (+1.340) \\
\bottomrule
\end{tabular}
}
}
\end{table}

\subsection{Compare with T2V}

While results in main paper demonstrates promising gains from joint generation over the baseline, we further investigate whether joint generation provides advantages over standard supervised fine-tuning (SFT) for text-to-video generation. We conduct extensive ablation studies across different modalities, training models with identical data and steps to ensure fair comparison of their text-to-video capabilities.

As shown in Table~\ref{tab_comparison_modalities}, all modality configurations with joint generation achieve significant improvements over both the baseline and T2V-only training. Each auxiliary modality contributes distinct supervisory signals that enhance the model's visual understanding, confirming the complementary nature of different modalities. Moreover, unified multi-modal training outperforms single-modality joint training by achieving better balance across evaluation dimensions, with substantial gains in overall performance (Averaged column). These results validate that diverse modality supervision collectively strengthens video generation through mutual reinforcement rather than simply additive improvements.

\subsection{Scalability with Increasing Modalities}

To demonstrate UnityVideo's ability to continuously improve with expanded modality training, we evaluate performance scaling on both joint generation and controllable generation tasks. As shown in Table~\ref{tab_scalability_comparison}, UnityVideo achieves consistent performance gains across all metrics as the number of modalities increases. Specifically, we compare models trained with three modalities (depth, optical flow, and DensePose) against those trained with five modalities (additionally incorporating skeleton and segmentation).

The results reveal monotonic improvements across all evaluation criteria, confirming that our framework effectively leverages additional modality supervision without suffering from negative interference. This strong scalability suggests that UnityVideo's architecture can accommodate further expansion in both model parameters and modality diversity, potentially enabling emergent world perception capabilities as the framework scales. The consistent gains validate our unified training paradigm as a promising foundation for developing increasingly comprehensive video world models through continued modality integration.

\begin{table}[t]
\centering
\caption{
Analysis of the benefits brought by extended modal training for joint generation and control generation.
}
\vspace{-2mm}
{
\scriptsize
\label{tab_scalability_comparison}
\resizebox{1\linewidth}{!}{
\begin{tabular}{l|cccc}
\toprule
& \textbf{Subject} & \textbf{Background} & \textbf{Temporal} & \textbf{Motion} \\
& \textbf{Consistency} & \textbf{Consistency} & \textbf{Flickering} & \textbf{Smoothness} \\
\midrule
Baseline & 96.51 & 96.06 & 98.73 & 99.30 \\
\midrule
\multicolumn{5}{c}{\textit{Joint Generation}} \\
\midrule
Depth & 96.53 & 95.58 & 98.45 & 99.28 \\
Three Modalities & \underline{98.01} & \underline{97.24} & \underline{99.10} & \underline{99.44} \\
Five Modalities & \textbf{98.31} & \textbf{97.54} & \textbf{99.35} & \textbf{99.54} \\
\midrule
\multicolumn{5}{c}{\textit{Control Generation}} \\
\midrule
Depth & 97.78 & 96.79 & 98.80 & 99.30 \\
Three Modalities & \underline{97.83} & \underline{96.86} & \underline{98.87} & \underline{99.33} \\
Five Modalities & \textbf{97.87} & \textbf{97.32} & \textbf{99.57} & \textbf{99.39} \\
\bottomrule
\end{tabular}
}
}
\end{table}

\begin{figure*}[h]
    \centering
    \vspace{-4mm}
    \includegraphics[width=1.0\linewidth]{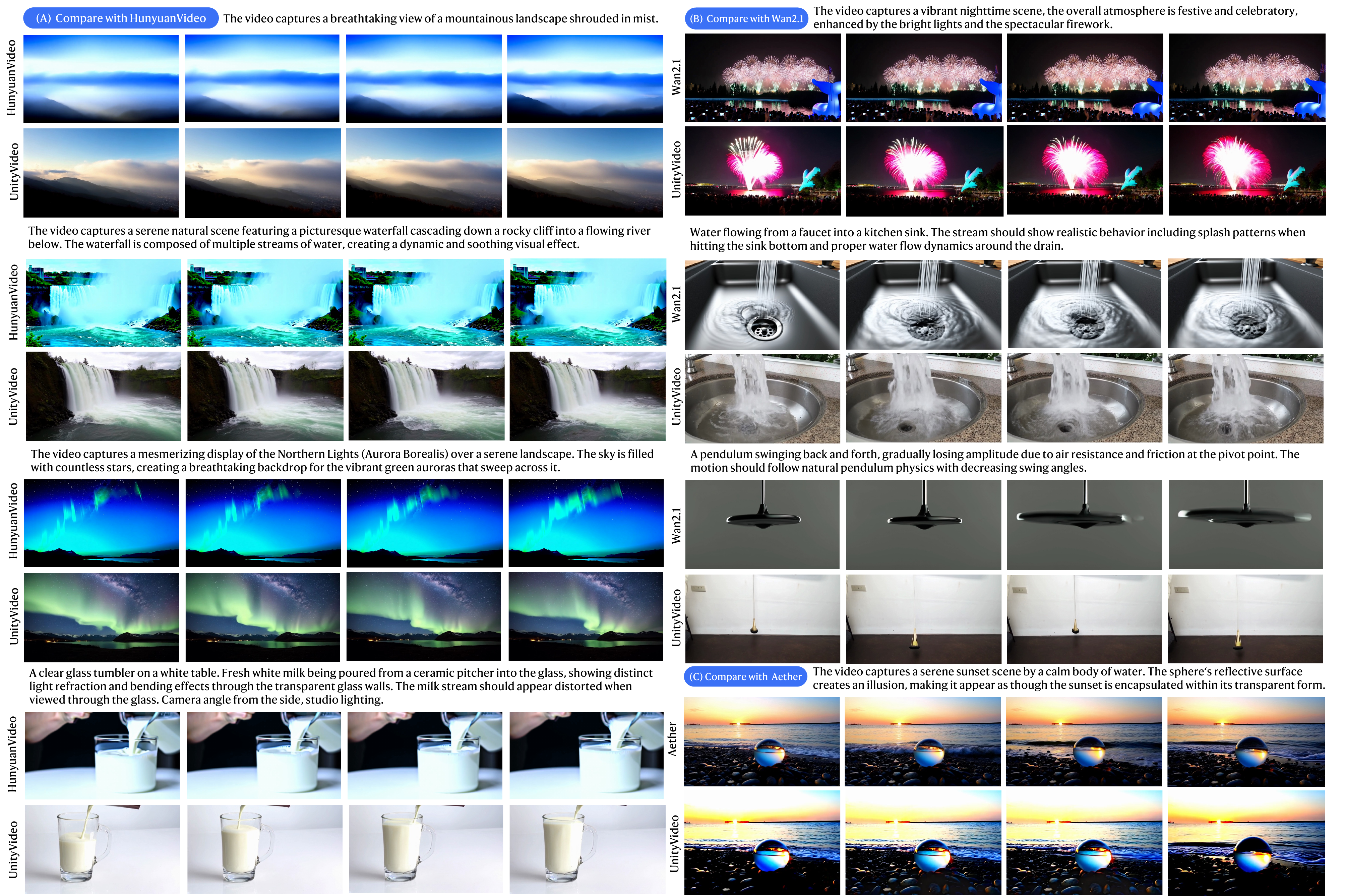}
    \caption{
    \textbf{Comparison of physical understanding.}
    UnityVideo demonstrates stronger physical reasoning and improved text alignment compared with current state-of-the-art video generation models.
 }
    \label{compare_all_base}
\end{figure*}

\subsection{The influence of different modalities}
As shown in main paper, incorporating additional modalities yields further improvements for the \textit{JointGeneration} task compared with training on a single modality. To examine whether this benefit also extends to \textit{ControlGeneration}, we conduct the ablation study summarized in Table~\ref{tab_control_comparison}. Here, \textit{Only} denotes models trained on ControlGeneration using a single modality, while \textit{Ours} refers to models trained jointly with three modalities. All training data and iteration budgets are kept strictly identical to ensure a fair comparison.

The results show that unified multimodal training consistently outperforms single-modality training on the ControlGeneration task. These findings demonstrate that \textit{UnityVideo} effectively strengthens positive cross-modal interactions across tasks, enabling each modality to benefit from the shared training paradigm.”

\begin{table}[t]
\centering
\caption{
The gain of joint modal training compared with single modal on ControlGeneration tasks.
}
\vspace{-2mm}
{
\scriptsize
\label{tab_control_comparison}
\resizebox{1\linewidth}{!}{
\begin{tabular}{l|ccccc}
\toprule
& \textbf{Subject} & \textbf{Background} & \textbf{Temporal} & \textbf{Motion} & \textbf{Averaged} \\
& \textbf{Consistency} & \textbf{Consistency} & \textbf{Flickering} & \textbf{Smoothness} & \\
\midrule
Baseline & 96.51 & 96.06 & 98.73 & 99.30 & 97.65 \\
\midrule
\multicolumn{6}{c}{\textit{Depth Modality}} \\
\midrule
ControlGen & 97.78 & 96.79 & 98.80 & 99.30 & 98.1675 \\
(Depth) & (+1.27) & (+0.73) & (+0.07) & (+0.00) & (+0.5175) \\
\textbf{Unified} & \textbf{97.83} & \textbf{96.86} & \textbf{98.87} & \textbf{99.33} & \textbf{98.2225} \\
(Depth) & (+1.32) & (+0.80) & (+0.14) & (+0.03) & (+0.5725) \\
\midrule
\multicolumn{6}{c}{\textit{Optical Flow Modality}} \\
\midrule
ControlGen & 97.40 & 96.59 & 98.67 & 99.23 & 97.9725 \\
(Optical Flow) & (+0.89) & (+0.53) & (-0.06) & (-0.07) & (+0.3225) \\
\textbf{ControlGen} & \textbf{97.47} & \textbf{96.72} & \textbf{98.83} & \textbf{99.32} & \textbf{98.0850} \\
(\textbf{Unified}) & (+0.96) & (+0.66) & (+0.10) & (+0.02) & (+0.4350) \\
\midrule
\multicolumn{6}{c}{\textit{Densepose Modality}} \\
\midrule
ControlGen & 97.01 & 96.47 & 98.58 & 99.10 & 97.790 \\
(Densepose) & (+0.50) & (+0.41) & (-0.15) & (+0.20) & (+0.5050) \\
{ControlGen} & \textbf{97.58} & \textbf{96.79} & \textbf{98.90} & \textbf{99.35} & \textbf{98.1550} \\
(\textbf{Unified}) & (+1.07) & (+0.73) & (+0.17) & (+0.05) & (+0.5050) \\
\bottomrule
\end{tabular}
}
}
\end{table}

\subsection{World perception comparison}

To further assess our model's world understanding capabilities, we conduct comprehensive evaluations using physics-focused prompts that test fundamental physical principles. As shown in Figure~\ref{compare_all_base}, we evaluate models on scenarios involving refraction, collision dynamics, and other physical phenomena that require accurate world modeling.

Our results demonstrate that UnityVideo exhibits superior understanding of physical laws compared to baseline methods. The model accurately captures light refraction through transparent media, realistic collision responses between objects, and physically plausible motion trajectories. These improvements stem from the complementary supervision provided by auxiliary modalities—depth enhances spatial reasoning, optical flow captures motion dynamics, and segmentation clarifies object boundaries—collectively enabling more accurate physical world modeling. This enhanced physical reasoning capability further validates the effectiveness of our unified multimodal training paradigm in developing world-aware video generation models.

\section{Details of OpenUni and UniBench}
\label{Appendix_details_of_OpenUni_and_UniBench}
\begin{figure*}[h]
    \centering
    \vspace{-8mm}
    \includegraphics[width=1.0\linewidth]{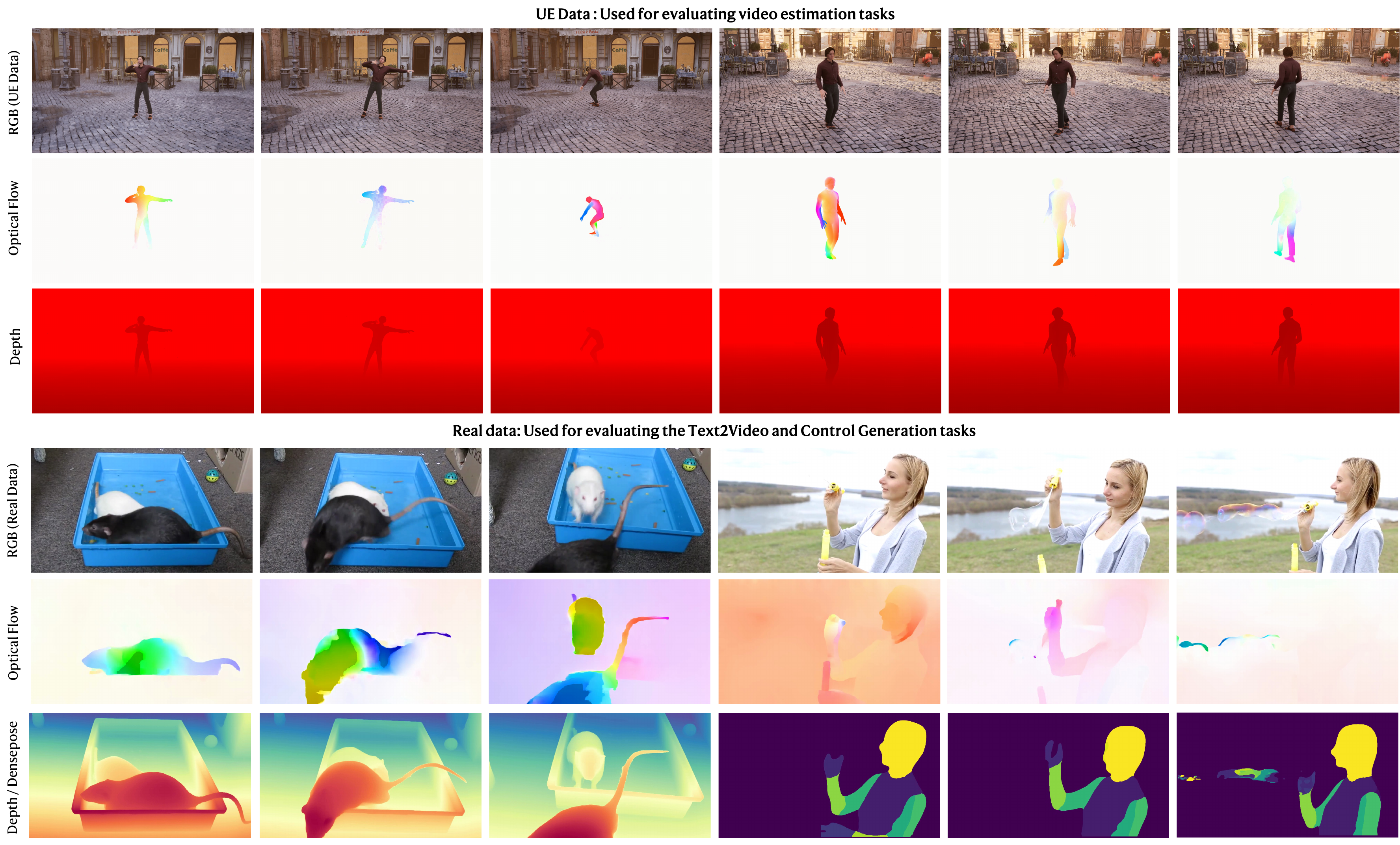}
    \caption{
    UniBench consists of two complementary components: (i) high-fidelity Unreal Engine depth data for evaluating depth estimation, and (ii) diverse real-world videos with rich multimodal annotations for assessing video generation quality.
 }
    \label{details_of_OpenUni_and_UniBench}
\end{figure*}

\subsection{OpenUni}
The OpenUni dataset leverages diverse data sources and comprehensive modality extraction to create a large-scale multimodal training corpus. We employ multiple pretrained models to extract modality-specific features and implement rigorous filtering pipelines to ensure data quality and usability.

Our data curation process follows strict quality criteria. We first filter source videos based on temporal, aesthetic, and resolution constraints: minimum duration of 5 seconds, aesthetic score exceeding 80/100, and spatial resolution above 512 pixels. Videos containing embedded text or subtitles are removed using OCR-based detection to prevent contamination of visual modalities. For each retained video, we extract corresponding modality annotations using specialized models—depth from Depth Anything V2, optical flow from RAFT, segmentation from SAM, skeleton from DWPose, and DensePose from Meta's implementation. Automated quality metrics further filter low-quality modality extractions, ensuring reliable ground-truth annotations across all modalities.

Through this systematic pipeline, we obtain approximately 1.3M high-quality multimodal video pairs, each containing synchronized annotations across five modalities. This comprehensive dataset enables effective unified training while maintaining consistency and quality across diverse visual representations.

\subsection{UniBench}

To address the absence of standardized evaluation benchmarks for unified multimodal video tasks, we construct UniBench with two distinct evaluation categories tailored to different task requirements. For video estimation tasks requiring ground-truth annotations, we generate synthetic data using Unreal Engine to obtain pixel-accurate depth maps and optical flow. As shown in Figure~\ref{details_of_OpenUni_and_UniBench}, for controllable generation and text-to-video tasks requiring diverse modality conditions, we curate high-quality samples from our test split.

Specifically, we create 200 synthetic video sequences with precise ground-truth depth and optical flow using Unreal Engine's rendering pipeline. These sequences feature significant camera and object motion to comprehensively evaluate depth estimation capabilities under challenging conditions. For generation tasks, we select 200 high-quality samples from the test subset, each containing complete annotations across all five modalities. This dual-track evaluation strategy enables rigorous assessment of both reconstruction accuracy and generation quality within our unified framework.

\section{More Visuals and Applications}
\label{Appendix_more_visuals_and_applications}

\begin{figure}[h]
    \centering
    \includegraphics[width=1.0\linewidth]{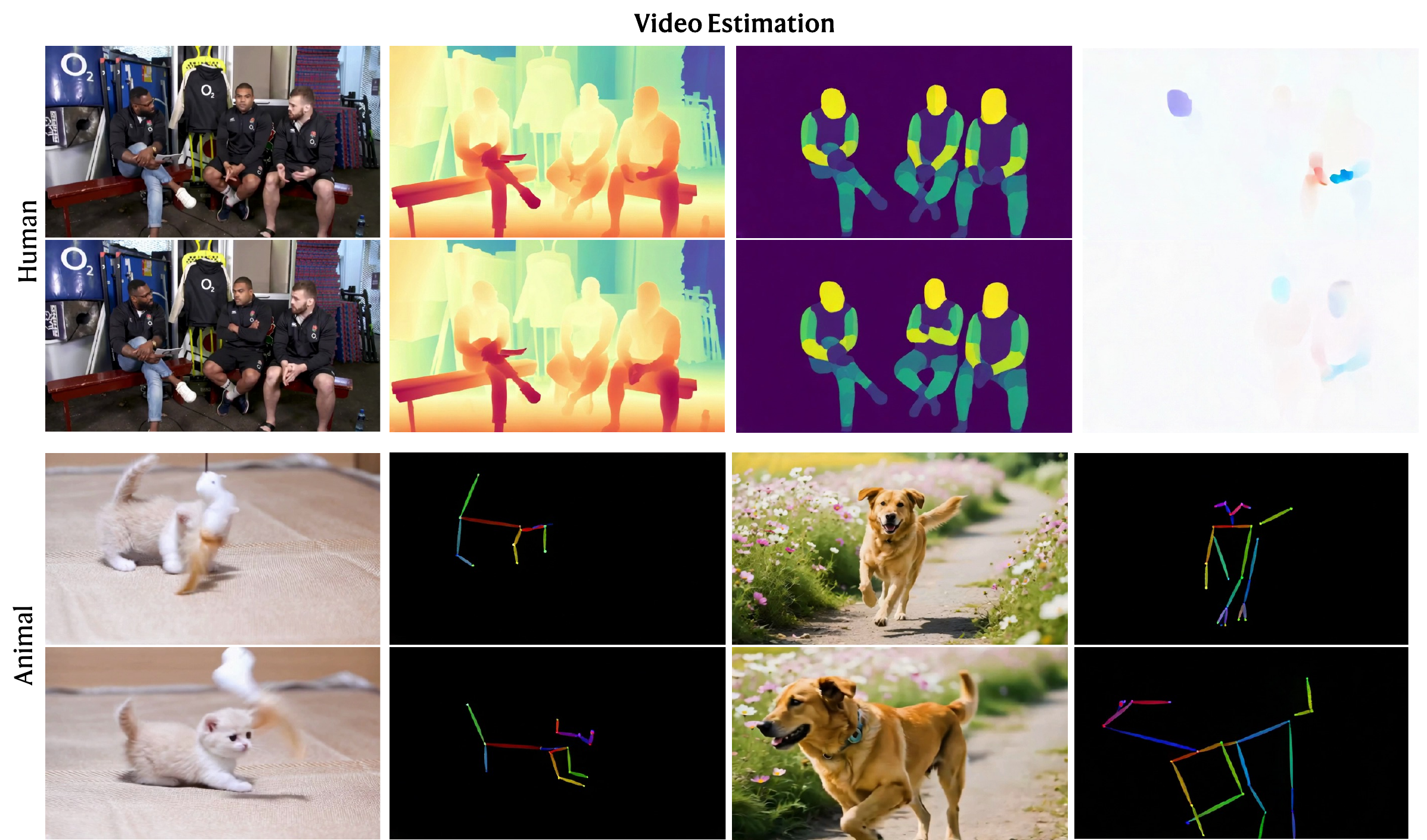}
    \caption{
    Representative outputs of UnityVideo on Video Estimation. The model consistently produces coherent RGB videos and aligned modalities—including densepose, optical flow, skeleton, and depth—demonstrating reliable cross-modal generation and estimation across diverse scenarios from human activities to animal motion.
    }
 \vspace{-4mm}
    \label{D_more_comparision_cases_2}
\end{figure}

\begin{figure}[h]
    \centering
    \vspace{-82mm}
    \includegraphics[width=1.0\linewidth]{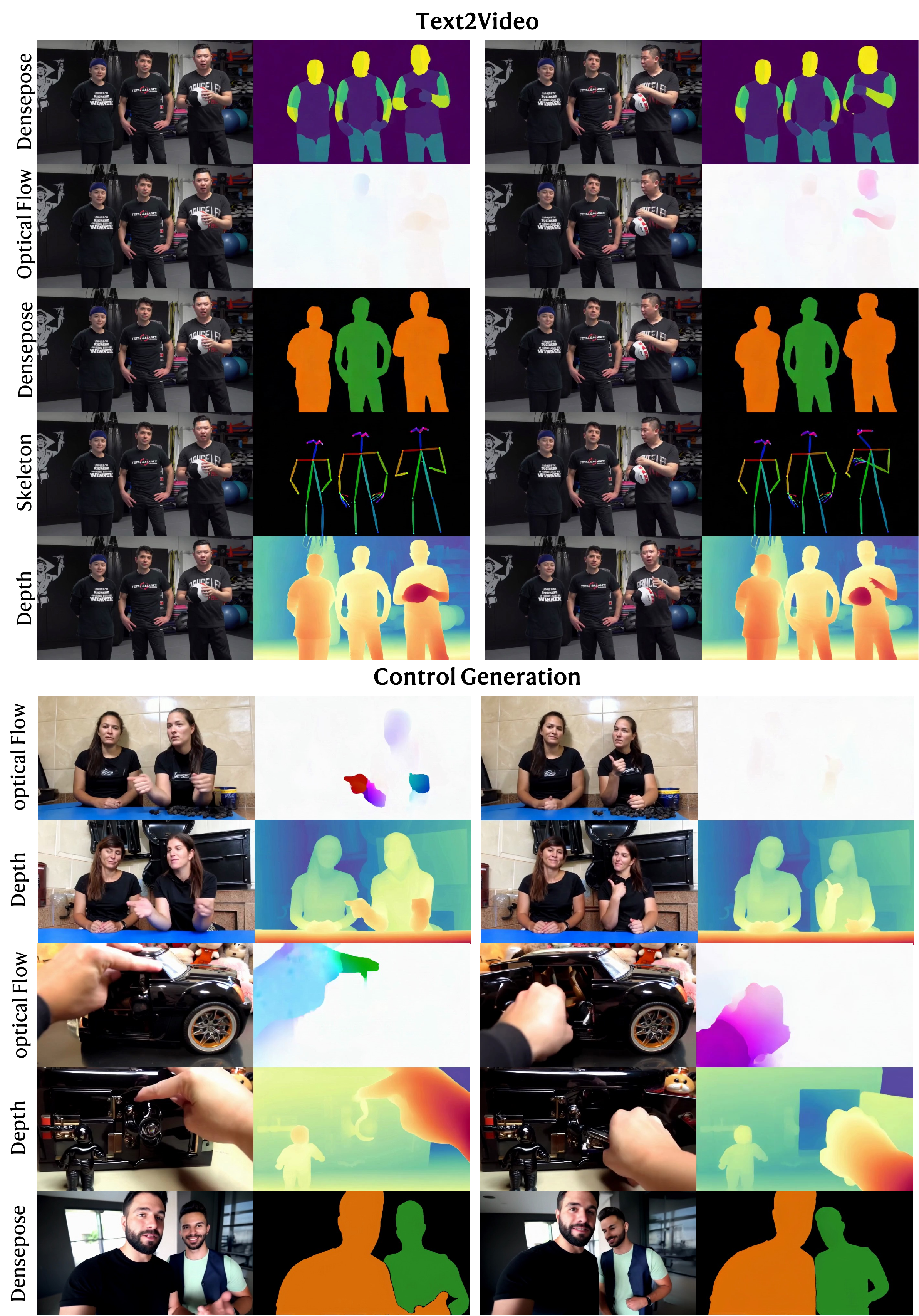}
    \caption{
    Representative outputs of UnityVideo on Text2Video and Control Generation. The model consistently produces coherent RGB videos and aligned modalities—including segmentation, densepose, optical flow, skeleton, and depth—demonstrating reliable cross-modal generation and estimation across various indoor and outdoor scenes with multiple subjects.
    }
 \vspace{-4mm}
    \label{D_more_comparision_cases_1}
\end{figure}

Figure~\ref{D_more_comparision_cases_2} and~\ref{D_more_comparision_cases_1} showcases UnityVideo's extensive generalization capabilities across three core tasks: controllable generation, video estimation, and joint generation. The model accepts arbitrary modality inputs for precise controllable generation while supporting flexible modality estimation for diverse subjects and scenarios.

Our framework demonstrates remarkable zero-shot generalization beyond its training distribution. While trained primarily on single-person data, UnityVideo successfully generalizes to multi-person scenarios for all modality estimations. Similarly, skeleton estimation capabilities trained on human subjects transfer effectively to animal motion capture without additional fine-tuning. The model also exhibits robust cross-domain transfer, accurately estimating depth and segmentation for out-of-distribution objects and scenes. These diverse examples collectively demonstrate that UnityVideo's unified training paradigm not only achieves technical integration across modalities but also develops genuine world understanding that enables flexible generalization to novel contexts and subjects.

\end{document}